\documentclass[sigconf]{acmart}

\AtBeginDocument{%
  \providecommand\BibTeX{{%
    \normalfont B\kern-0.5em{\scshape i\kern-0.25em b}\kern-0.8em\TeX}}}

\setcopyright{acmcopyright}
\copyrightyear{2023}
\acmYear{2023}
\acmDOI{10.1145/XXXXXX.XXXXXX}

\acmConference[SIGIR'23]{	
Proceedings of the 46th International ACM SIGIR Conference on Research and Development in Information Retrieval}{July 23--27, 2023}{Taipei, Taiwan, China.}
\acmBooktitle{Proceedings of the 46th International ACM SIGIR Conference on Research and Development in Information Retrieval (SIGIR '23), July 23--27, 2023, Taipei, Taiwan, China}
\acmPrice{15.00}
\acmISBN{978-1-4503-9408-6/23/07}

\begin{document}

\title{Friend Ranking in Online Games via Pre-training Edge Transformers}
\author{Liang Yao, Jiazhen Peng, Shenggong Ji, Qiang Liu, Hongyun Cai, Feng He, Xu Cheng}
\email{{dryao, brucejzpeng, shenggongji, pobliu, laineycai, fenghe, alexcheng}@tencent.com}
\affiliation{%
  \institution{Tencent Inc.}
  \city{Shenzhen}
  \country{China}
}

\renewcommand{\shortauthors}{Yao, et al.}

\begin{abstract}

Friend recall is an important way to improve Daily Active Users (DAU) in online games. The problem is to generate a proper lost friend ranking list essentially. Traditional friend recall methods focus on rules like friend intimacy or training a classifier for predicting lost players' return probability, but ignore feature information of (active) players and historical friend recall events. In this work, we treat friend recall as a link prediction problem and explore several link prediction methods which can use features of both active and lost players, as well as historical events. Furthermore, we propose a novel Edge Transformer model and pre-train the model via masked auto-encoders. Our method achieves state-of-the-art results in the offline experiments and online A/B Tests of three Tencent games.
\end{abstract}

\ccsdesc[500]{Computing methodologies~Neural networks}
\ccsdesc[500]{Information systems~Social networks}

\keywords{Friend Ranking, Link Prediction, Transformer, Pre-training}



\maketitle

\section{Introduction}

\begin{figure*}[t]
  \centering
  \includegraphics[width = 0.84 \textwidth]{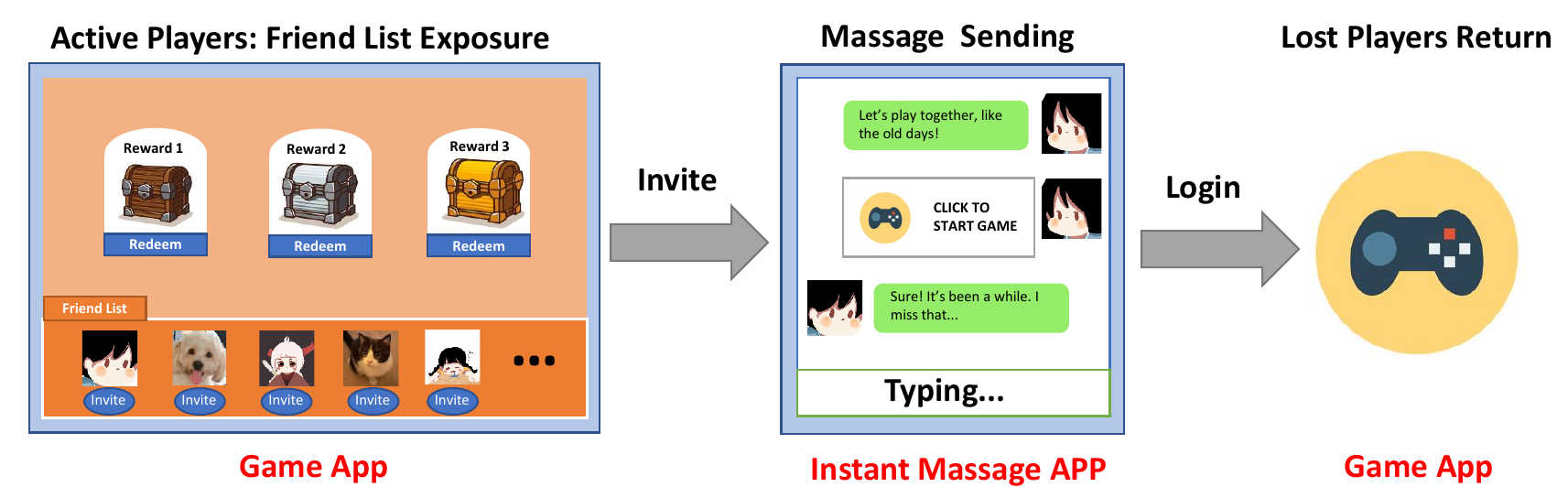}
  \caption{Illustrations of friend recall in online games. On the left, when an active player (say X) logs in the Game A, X can see a list of friends (in the bottom rectangle)  who have been inactive in the game recently. X can click and invite them to return to the game and play with X together. In the middle, X invited a friend (say Y) in the friend list and sent Y a link to the game. On the right, Y saw the invitation and clicked the link, Y then returned to the game. The above process is a successful story, X may not invite any friend, and Y may not accept the invitation. }
  \label{fig:friend_recall}
\end{figure*}

Online gaming is one of the largest industries on the Internet, generating tens of billions of dollars in revenues annually.
Tencent is China's largest Internet company and the largest game service provider. It has more than 800 million gaming users\footnote{https://www.tencent.com/en-us/articles/2200928.html}. The payment revenue from games accounts for 31$\%$ of Tencent’s overall revenue\footnote{https://www.statista.com/statistics/527280/tencent-annual-online-games-revenue/}.

In online games, one critical problem is how to design strategies to keep players playing. Friend recall is an essential way for the purpose. It means letting active players in the game invite their lost friends back. An illustration of friend recall is in Figure~\ref{fig:friend_recall}. An active player X in the game can see a number of events. In some events, the player can earn awards (e.g., coins, hero skins) by inviting his/her lost friend. X can click a photo of his friend Y in the recommendation list, then X will send an instant message to the clicked friend. The friend Y may return to the game after seeing the message. If Y is back, X will earn more rewards. 

The friend recall problem is to generate a proper lost friend list essentially. There are two main traditional methods in our practice. One is to sort the lost friends by intimacy scores. A lost friend Y with a higher intimacy score will be listed before another friend Z with a lower intimacy score. The intimacy score between player X and Y is calculated by summing the interaction activities in the game (e.g., giving a gift, or playing in the same room). The basic idea is that intimate friends are more likely to be invited, and an invited friend is more likely to be back. The second is training a classifier for predicting lost players’ return probability. The method collects features of lost players and uses natural return activities as labels. If a lost player logs in the game without invitation, the label is positive, otherwise, negative. A lost player with a higher natural return probability will be given a higher rank. The idea is to give players who are more likely to be back more chances to be invited. 

The above two methods have some major limitations: 1). The intimacy scores neglect both features of active and lost players, an active player may not be willing to invite, and a lost friend may not be willing to return. 2). The natural return classifier only uses features of lost players, but the features of active players are ignored. 3). Both methods could not utilize information from historical friend recall events. Historical events indicate which active players (with features before the event) invited their friends, and which lost players are invited and then returned. 

In this work, we overcome the above limitations by treating friend recall as a link prediction problem. The problem is to predict the edge existing probability in a bipartite graph. The bipartite graph contains two kinds of nodes: active players and lost players. The edge is labeled as positive if an active player successfully invites a lost player back in a game event. We explore several popular link prediction 
and friend ranking models for the task. Furthermore, we propose a novel Edge Transformer model and pre-train the model via Masked Auto-Encoders (MAE). The proposed method outperforms existing models and the two traditional methods in both offline and online evaluation. The implementation of our proposed method is available at:~\url{https://github.com/yao8839836/edge_mae}. To summarize, our contributions are as follows:

\begin{itemize}
    \item To the best of our knowledge, this is the first study to systematically investigate the real-world friend recall problem.
    \item We propose a new Edge Transformer model and improve it by pre-training with masked auto-encoders and massive unlabeled edges. The model outperforms state-of-the-art link prediction models in our task.
\end{itemize}

\section{Data}

We built our datasets from past events of Game A and Game B. The real game names are hidden for privacy reasons. In past events, an active player X who clicked at least one of his/her lost friends and X's lost friends are collected as labeled edges and added to the training and validation dataset. If X clicked a lost friend Y, and Y was back, the edge is labeled as positive. If Y is not back, the edge is labeled as negative. All edges in the training and validation dataset without an invitation are also labeled as negative. For pre-training, we use massive unlabeled edges. All active players who did not invite any friends together with their lost friends are collected as unlabeled edges for pre-training. We construct node features, i.e., gaming activities and payment information, for two kinds of players and take intimacy scores as edge features. The statistics of the two datasets are listed in Table~\ref{tab:data}.

\section{Method}

\begin{table}[t]
  \caption{Data Statistics.}
  \label{tab:data}
  \begin{tabular}{ccc}
    \toprule
    Dataset & Game A& Game B \\
    \midrule
    \# labeled edges & 6,396,834 & 11,982,659\\
    \# train edges & 5,117,125 & 9,591,653\\
    \# val edges & 1,279,709 & 2,391,006\\
    \# unlabeled edges & 24,946,624 & 3,422,556,670\\
    \# active players features dim & 80 & 25\\
    \# lost players features dim & 80 & 26\\
    \# edge features dim & 4 & 1\\
  \bottomrule
\end{tabular}
\end{table}

\begin{figure}[t]
  \centering
  \includegraphics[width = 0.27 \textwidth]{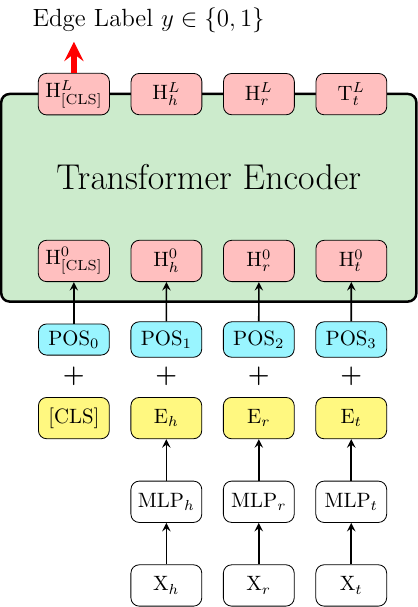}
  \caption{The proposed Edge Transformer model. For an edge, we linearly embed the head node feature $X_h$, edge feature $X_r$, and tail node feature $X_t$ to the same dimension via three MLPs, then add position embeddings and feed the resulting vectors to a standard Transformer encoder. An extra learnable “classification token” is added to the first of the sequence to perform classification. }
  \label{fig:edge_transformer}
\end{figure}

\begin{figure}[t]
  \centering
  \includegraphics[width = 0.42 \textwidth]{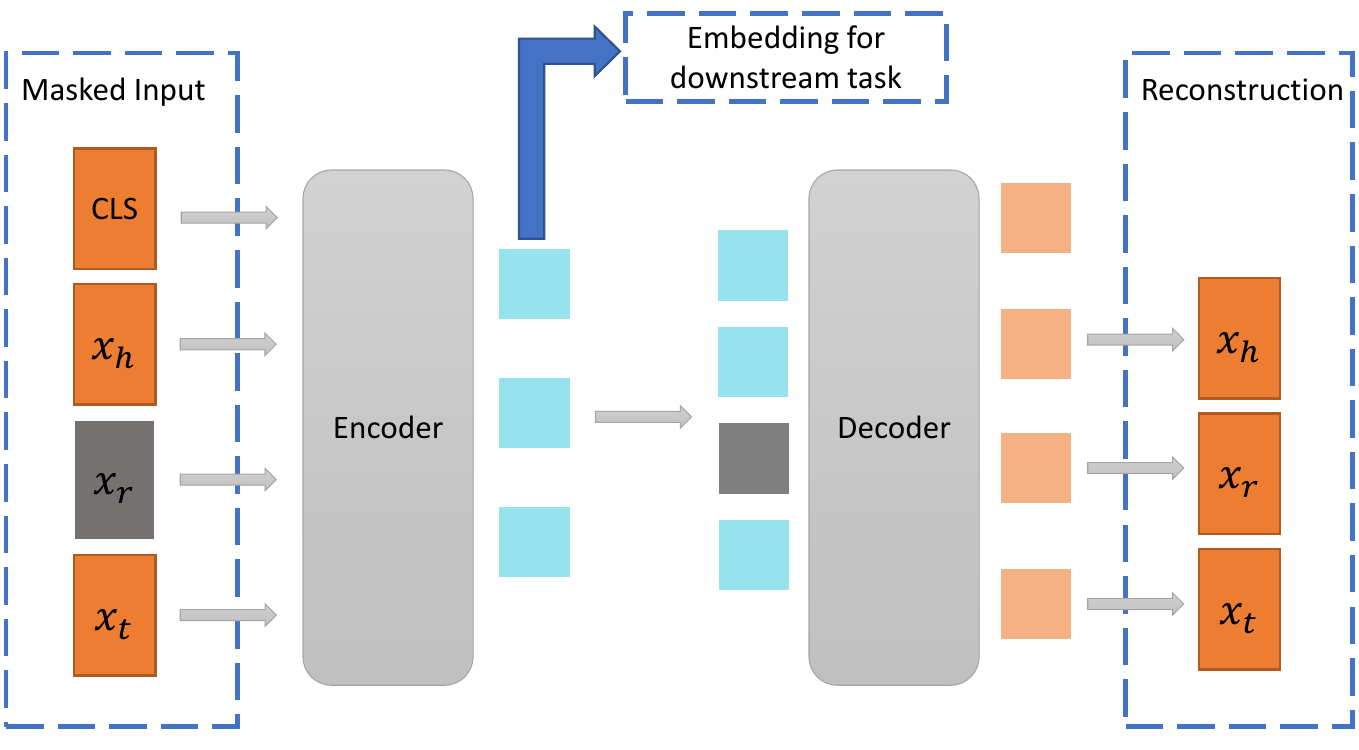}
  \caption{The proposed Edge MAE model. During pre-training, a 
random subset of edge tokens (e.g., $1/3$) is masked out. The
encoder is performed on the subset of visible tokens. Masked
tokens are introduced after the encoder, and the full set of encoded tokens and masked tokens are processed by a small decoder
that reconstructs the original features of nodes or edges. After pre-training, the decoder is discarded and the encoder is applied to uncorrupted 
 edges (full sets of tokens) for edge classification.}
  \label{fig:edge_mae}
\end{figure}

Inspired by the Transformer~\cite{vaswani2017attention} scaling successes in natural language processing (NLP)~\cite{devlin2019bert,brown2020language} and computer vision (CV)~\cite{dosovitskiy2020image}, we experiment with applying a standard Transformer directly to edges, with the fewest possible modifications, as the scalable NLP and CV Transformer architectures and their implementations can be reused. We treat node and edge features as word tokens in NLP, and image patches in CV. We take the sequence of linear embeddings of these features as an input to a Transformer.

An overview of the Edge Transformer model is depicted in Figure~\ref{fig:edge_transformer}. The standard Transformer receives  a 1D sequence of token embeddings as input. To handle edges, we linearly embed the head node feature $X_h$ (for active players), edge feature $X_r$ (intimacy scores), and tail node feature $X_t$ (for lost players) to the same dimension via three multilayer perceptions (MLP). The Transformer uses latent vector size $D$ through all of its layers, so we map the features to $D$ dimensions. We refer to the output of these projections as the token embeddings. 

Similar to BERT~\cite{devlin2019bert} and ViT~\cite{dosovitskiy2020image}'s [CLS] token, we add a learnable embedding to the sequence of embedded tokens. Its state at the output of the Transformer encoder H$_{\text{[CLS]}}^L$ serves as the edge representation for classification. Position embeddings POS$_0$, POS$_1$, POS$_2$ and POS$_3$ are added to the tokens embeddings to keep positional information. The resulting
sequence of embedding vectors H$_{\text{[CLS]}}^0$, H$_h^0$, H$_r^0$ and H$_t^0$ is as input to the encoder. The details of the Transformer encoder are the same as in~\cite{vaswani2017attention}.

The Transformer encoder can be only trained on labeled edges, while massive unlabeled edges are not used. Inspired by MAE~\cite{he2022masked} in CV which pre-trains masked auto-encoders on unlabeled images, we propose to improve Edge Transformer with an Edge MAE model. The model is depicted in Figure~\ref{fig:edge_mae}. Following MAE, Edge MAE randomly masks a proportion of input tokens. The encoder of an Edge MAE is an Edge Transformer but only applied on unmasked tokens. All four edge tokens are the input to the Edge MAE decoder. The decoder is another series of Transformer blocks and is only used in pre-training to perform node/edge feature reconstruction tasks. The reconstruction target is to predict each feature value of mask tokens. The last layer of the decoder is a linear projection that maps the hidden vector (size $D$) to the original feature dimension. Our loss function
computes the mean squared error (MSE) between the reconstructed and original features. After pre-training, the parameters of the encoder are used as the initialization of the Edge Transformer model for edge classification fine-tuning.  

\begin{table*}[t]
\footnotesize
  \caption{Ranking performances of different methods on validation data of Game A. The best result is in bold font. We run all models 10 times and found Edge MAE significantly outperforms baselines based on student $t$-test ($p < 0.05$).} 
  \label{tab:GameA_val}
  \begin{tabular}{c|cccccccc}
    \toprule
     Method & Hits@1 & Hits@3 & Hits@5 & Hits@10 & MR & MRR & \#  top 5 back & \# top 10 back \\
    \midrule
     Intimacy & 0.2804	& 0.5857 &	0.8116 & 0.9368	& 4.0638 &	0.4824	& 44,957 &	51,897 \\
    XGB for lost players &0.5305 &	0.8498 &	0.9429	& 0.9856 &	2.2507&	0.6993&	52,235&	54,598\\
    
     Edge MLP &0.5482	& 0.8616&	0.9491 &	0.9871&	2.1490&	0.7130&	52,576&	54,682\\
     Edge XGB &0.5492 &	0.8622	& 0.9487	&0.9872	& 2.1439	&0.7129	&52,554	&54,703\\
     Bilinear &0.5478	& 0.8625 &	0.9490	& 0.9874&	2.1491&	0.7128	& 52,569	& 54,697\\
     DistMult &0.5424 &	0.8593	& 0.9475	& 0.9871 &	2.1751 &	0.7088	& 52,488	& 54,673\\
     TransE &0.5276 &	0.8460	& 0.9419	& 0.9839	& 2.2715	& 0.6971 &	52,179	&54,505\\     
     ConvKB &0.5478&	0.8624&	0.9494&	0.9872	&2.1462	& 0.7130 &	52,594&	54,688\\
     TranS &0.5450 &  0.8616	&0.9477	 &	0.9874	& 2.1628 &	0.7108	&52,499 	& 54,696\\  
     Pairwise Ranking & 0.5353& 0.8564	&	0.9467 &	0.9862	& 2.1982 &	 0.7040	& 52,443	&54,630 \\      
     Edge Transformer&0.5475	& 0.8627 &	0.9488&	\textbf{0.9875}&	2.1470&	0.7127&	52,562&	\textbf{54,704}\\
     Edge MAE & \textbf{0.5497} &	\textbf{0.8628} &	\textbf{0.9495} &	\textbf{0.9875}	&\textbf{2.1437}&	\textbf{0.7139} &	\textbf{52,596} &	\textbf{54,704}\\
    \bottomrule
  \end{tabular}
\end{table*}

\begin{table*}[t]
\footnotesize
  \caption{Ranking performances of different methods on validation data of Game B. The best result is in bold font. We run all models 10 times and found Edge MAE significantly outperforms baselines based on student $t$-test ($p < 0.05$).}
  \label{tab:GameB_val}
  \begin{tabular}{c|cccccccc}
    \toprule
     Method & Hits@1 & Hits@3 & Hits@5 & Hits@10 & MR & MRR & \# top 5 back & \# top 10 back \\
    \midrule
     Intimacy & 0.1324 &	0.2933	&0.4661	&0.6238	&12.6622 &	0.2779&	61,461	& 82,249 \\
    XGB for lost players &0.4055 &	0.6582 &	0.774 &	0.9011 &	4.1478	& 0.5659 &	102,052&	118,811\\
    
     Edge MLP &0.4428	& 0.699	& 0.8052	& 0.9148	& 3.7798 &	0.5998	& 106,161 &	120,616\\
     Edge XGB &0.4399 &	0.6994	& 0.8054 &	0.9157	& 3.7906 &	0.5981 &	106,197	&120,742\\
     Bilinear &0.4435 &	0.6998	& 0.8047&	0.9154	& 3.7788 & 0.6003	& 106,103&	120,697\\
     DistMult &0.4149&	0.6709	& 0.7856	& 0.9067 &	4.0437 &	0.5755	& 103,583	& 119,555\\
     TransE &0.1502 &	0.3582	& 0.5131	&0.6814&	10.7953	&0.3120	&67,651	& 89,848\\     
     ConvKB &0.4424 &	0.6982 &	0.8054 &	0.9155	& 3.7845 &	0.5993	& 106,190	& 120,704\\
     TranS &0.4335 &	0.6906	 &0.7992	&	 0.9134&  3.8723 &	0.5919	&105,382 	& 120,439\\  
     Pairwise Ranking &0.3653 & 0.6158	&0.7360 &	0.8771 &	4.7332	& 0.5291 &	97,039 &115,651\\     
     Edge Transformer& 0.4441 &	0.7004	& 0.8068 &	0.9159 &	3.7630	& 0.6011 &	106,379 &	120,761\\
     Edge MAE & \textbf{0.4451} &	\textbf{0.7023} &	\textbf{0.8091} &	\textbf{0.9173}	&\textbf{3.7485}&	\textbf{0.6021} &	\textbf{106,675} &	\textbf{120,953}\\
    \bottomrule
  \end{tabular}
\end{table*}

\begin{table}
\footnotesize
  \caption{Online A/B test result of different methods in a recent event of Game A.}
  \label{tab:onlineA}
  \begin{tabular}{ccccc}
    \toprule
    Method & Click Rate & Return Rate & Conversion Rate &  Improvements\\
    \midrule
    XGB for lost& 4.43 $\%$ & 13.70 $\%$ & 1.2556 $\%$ & --\\
    Bilinear & 4.46 $\%$ & 14.37 $\%$  & 1.3467 $\%$ & 7.26$\%$\\
    ConvKB & 4.43 $\%$ &  14.61 $\%$ & 1.3378 $\%$ &6.55$\%$\\
    Edge MAE & \textbf{4.65 $\%$} & \textbf{15.24 $\%$} & \textbf{1.4704 $\%$} & \textbf{17.11 $\%$}\\
  \bottomrule
\end{tabular}
\end{table}

\begin{table}
\footnotesize
  \caption{Online A/B test result of different methods in a recent event of Game C.} 
  \label{tab:onlineB}
  \begin{tabular}{ccccc}
    \toprule
    Method & Click Rate & Return Rate & Conversion Rate & Improvements\\
    \midrule
    XGB for lost& 8.41 $\%$ & 26.55 $\%$ & 7.2168 $\%$ & --\\
    ConvKB & 9.41 $\%$ & 26.59 $\%$ & 7.6109 $\%$ & 5.46$\%$ \\
    Edge MAE & \textbf{9.64 $\%$} & \textbf{27.86 $\%$ }&  \textbf{8.1204 $\%$} & \textbf{12.52 $\%$}\\
  \bottomrule
\end{tabular}
\end{table}

\section{Evaluation}
In this section, we evaluate our Edge Transformer and Edge MAE in two settings. Specifically, we want to determine:
\begin{itemize}
    \item Can our model achieve satisfactory ranking performance on validation data?
    \item Can our model outperform traditional methods in online A/B tests?
\end{itemize}

\subsection{Baselines}

We compare our proposed method with the two traditional methods Intimacy scores and XGBoost~\cite{chen2016xgboost} for lost players, as well as several popular link prediction models: TranS~\cite{zhang2022trans}, ConvKB~\cite{nguyen2018novel}, TransE~\cite{bordes2013translating}, DistMult~\cite{yang2015embedding} and Bilinear\footnote{\url{https://pytorch.org/docs/stable/generated/torch.nn.Bilinear.html}}. We also perform MLP and XGB on the concatenation of three features $X_h$, $X_r$ and $X_t$ and we call them Edge MLP and Edge XGB. Additionally, we compare our method with the pairwise ranking in a recent friend ranking method GraFRank~\cite{sankar2021graph}.

\subsection{Parameter Settings}

For Edge Transformer and Edge MAE, we set the latent vector size $D$ as 256, the dropout rate as 0.0, and the number of attention heads as 3. We tuned the encoder layer as 6 for Game A and 2 for Game B. The decoder layer of Edge MAE is tuned as 1. For pre-training Edge MAE, we tuned the learning rate as 1.5e-4, weight decay as 0.05, the mask ratio as $1/3$, and batch size as 2048. For fine-tuning, we tuned the learning rate as 0.001, and the batch size as 2048. The model was pre-trained with 20 epochs (about 15.5 hours for Game A and 153 hours for Game B) and fine-tuned for 50 epochs (about 3.2 hours for Game A and 4.5 hours for Game B) on an NVIDIA A100 GPU. For baseline methods, we use default parameter settings in their original papers or implementations. We found small changes of parameters for our method and baselines didn't change the results much. For a fair comparison, we add three MLP modules to produce the token embeddings $E_h$, $E_r$, and $E_t$ as the initialization before the input of link prediction and pairwise ranking models.

\subsection{Validation Results}

The ranking performances of different methods on the validation set are listed in Table~\ref{tab:GameA_val} and Table~\ref{tab:GameB_val}. We use the commonly used ranking metrics Hits@k, Mean Rank (MR, lower is better), Mean Reciprocal Rank (MRR), and the number of players who were back in a method's top 5/10 recommendation. From the two tables, we can see that, Edge Transformer outperforms almost all baseline models, and the results can be further improved by pre-training Edge MAE on unlabeled edges. The encouraging results showcase the effectiveness of the Transformer model and masked pre-training with massive unlabeled data. For more in-depth performance analysis, we found link prediction models are much better than the two traditional methods: even a classical MLP or XGB can perform quite well, which indicates link prediction is a proper setting for the friend recall task. The main reasons why our proposed method outperforms others are 1). Self-attention allows Edge Transformer to integrate information across nodes and edges. 2). The reconstruction target in MAE allows the model to learn the prior distribution of node and edge features.

\subsection{Online A/B Tests}

We conducted online A/B tests in recent events of two games. We randomly assigned different algorithm labels to each active player and his/her lost friends, the lost friends are ranked by corresponding models. The statistics are calculated after the online events given in Table~\ref{tab:onlineA} and Table~\ref{tab:onlineB}. The click rate is the proportion of active players who invited at least one friend after seeing the friend list. The return rate is the proportion of players who went back after being invited. The conversion rate (the final metric) is the number of return players divided by the number of active players who saw the friend list. We can see the proposed Edge MAE achieves the best results, and link prediction models perform better as they model the whole return process entirely.

\section{Related Work}

Transformers~\cite{vaswani2017attention} was first introduced for machine translation, then become the de-facto standard for NLP tasks~\cite{devlin2019bert}. Recently, Transformers became dominant in CV since the invention of ViT~\cite{dosovitskiy2020image}. The ViT models are further improved by pre-training masked auto-encoders on unlabeled images~\cite{he2022masked}. Transformers have also been explored in the graph domain~\cite{min2022transformer}. Our method is inspired by the line of works, the distinction is we train a Transformer directly on edges (features) while most existing graph Transformer architectures are trained on graph-level or node-level tasks. A recent work~\cite{bergen2021systematic} applies Transformers to edges in complete graphs for NLP tasks, but the model is not pre-trained.

With a variety of real-world applications, link prediction has been recognized as of great importance and attracted the wide attention of the research community. Existing link prediction approaches can be categorized into three families: heuristic feature-based, embedding-based and neural network-based methods. The closest line of works to ours in link prediction is the knowledge graph embedding approach~\cite{wang2017knowledge} which learns node and edge embeddings for triples. The major difference is that in knowledge graph embeddings, the node/edge embeddings are randomly initialized, while we explicitly use features of nodes and edges.

There are also a number of studies for friend ranking~\cite{sankar2021graph} or churn prediction in social platforms~\cite{oskarsdottir2017social,verbeke2014social,richter2010predicting}. These works focus on building network features for a user, then feeding user features to a classifier (like XGB for lost players). In contrast, our method learns representation for an edge (a pair of users with interactions) in an end-to-end manner.

\section{Conclusion}

In this work, we study the real-world friend recall problem and solve the problem via a novel Edge Transformer model with masked auto-encoders. The method outperforms traditional strategies, link prediction, and ranking models. We plan to improve the model with more unlabeled edges and multi-modal information for future work.


\bibliographystyle{ACM-Reference-Format}
\bibliography{dryao}


\end{document}